\documentclass{article}
\usepackage{ijcai20}

\usepackage{amsmath,amsfonts,bm}

\def\eqref#1{equation~\ref{#1}}

\def\1{\bm{1}}

\DeclareMathAlphabet{\mathsfit}{\encodingdefault}{\sfdefault}{m}{sl}
\SetMathAlphabet{\mathsfit}{bold}{\encodingdefault}{\sfdefault}{bx}{n}

\DeclareMathOperator{\sign}{sign}

\usepackage{times}
\usepackage{soul}
\usepackage{url}
\usepackage[hidelinks]{hyperref}
\usepackage[utf8]{inputenc}
\usepackage[small]{caption}
\usepackage{graphicx}
\usepackage{amsmath}
\usepackage{amsthm}
\usepackage{booktabs}
\usepackage[]{threeparttable}
\usepackage{multirow}
\usepackage{enumitem}
\usepackage[linesnumbered,ruled,vlined]{algorithm2e}
\usepackage[aboveskip=0pt,belowskip=-2pt]{subcaption}
\usepackage{pifont}
\usepackage{xcolor}
\newcommand\red[1]{{\color{black}#1}}

\title{Direct Quantization for Training Highly Accurate Low Bit-width\\ Deep Neural Networks}

\author{
Tuan Hoang$^{1}$  \and
Thanh-Toan Do$^2$\and
Tam V. Nguyen$^3$ \textnormal{and}
Ngai-Man Cheung$^{1}$
\affiliations
$^1$Singapore University of Technology and Design\\
$^2$University of Liverpool\\ 
$^3$University of Dayton\\
\emails
nguyenanhtuan\_hoang@mymail.sutd.edu.sg, thanh-toan.do@liverpool.ac.uk \\ tamnguyen@udayton.edu, ngaiman\_cheung@sutd.edu.sg \\
}

\begin{document}
\newcommand{\etal}{{\textit{et al. }}}

\newcommand{\ba}{\mathbf{a}}
\newcommand{\bb}{\mathbf{b}}
\newcommand{\bc}{\mathbf{c}}
\newcommand{\bk}{\mathbf{k}}
\newcommand{\bo}{\mathbf{o}}
\newcommand{\bp}{\mathbf{p}}
\newcommand{\bq}{\mathbf{q}}
\newcommand{\bs}{\mathbf{s}}
\newcommand{\bt}{\mathbf{t}}
\newcommand{\bu}{\mathbf{u}}
\newcommand{\bv}{\mathbf{v}}
\newcommand{\bw}{\mathbf{w}}
\newcommand{\bx}{\mathbf{x}}
\newcommand{\by}{\mathbf{y}}
\newcommand{\bz}{\mathbf{z}}

\newcommand{\bA}{\mathbf{A}}
\newcommand{\bB}{\mathbf{B}}
\newcommand{\bC}{\mathbf{C}}
\newcommand{\bD}{\mathbf{D}}
\newcommand{\bE}{\mathbf{E}}
\newcommand{\bF}{\mathbf{F}}
\newcommand{\bG}{\mathbf{G}}
\newcommand{\bI}{\mathbf{I}}
\newcommand{\bL}{\mathbf{L}}
\newcommand{\bO}{\mathbf{O}}
\newcommand{\bR}{\mathbf{R}}
\newcommand{\bS}{\mathbf{S}}
\newcommand{\bT}{\mathbf{T}}
\newcommand{\bU}{\mathbf{U}}
\newcommand{\bV}{\mathbf{V}}
\newcommand{\bW}{\mathbf{W}}
\newcommand{\bX}{\mathbf{X}}
\newcommand{\bY}{\mathbf{Y}}
\newcommand{\bZ}{\mathbf{Z}}

\newcommand{\hx}{\hat{x}}
\newcommand{\hbW}{\hat{\mathbf{W}}}

\newcommand{\bOnes}{\mathbf{1}}
\newcommand{\bZeros}{\mathbf{0}}

\newcommand{\bTheta}{\mathbf{\Theta}}

\newcommand{\bLm}{\mathbf{L}^m}

\newcommand{\transpose}{\hspace{-0.15em}^\top\hspace{-0.15em}}
\newcommand{\eq}{\hspace{-0.15em}=\hspace{-0.15em}}

\newcommand{\bbR}{\mathbb{R}}

\newcommand{\ndis}{\mathcal{N}}
\newcommand{\xset}{\mathcal{X}}
\newcommand{\oset}{\mathcal{O}}
\newcommand{\vset}{\mathcal{V}}
\newcommand{\tset}{\mathcal{T}}
\newcommand{\uset}{\mathcal{U}}
\newcommand{\dis}{\mathcal{D}}

\newcommand{\func}{\boldsymbol{f}}
\newcommand{\m}{{(m)}}
\newcommand{\mt}{{(t)}}
\newcommand{\M}{{(M)}}

\newcommand{\round}{\mathrm{round}}

\maketitle

\begin{abstract}
This paper proposes two novel techniques to train deep convolutional neural networks with low bit-width weights and activations. First, to obtain low bit-width weights, most existing methods obtain the quantized weights by performing  quantization on the full-precision network weights. However, this approach would result in some mismatch: the gradient descent updates full-precision weights, but it does not update the quantized weights. To address this issue, we propose a novel method that enables {direct} updating of quantized weights {with learnable quantization levels} to minimize the cost function using gradient descent. Second, to obtain low bit-width activations, existing works consider all channels equally. However, the activation quantizers could be biased toward a few channels with high-variance. To address this issue, we propose a method to take into account the quantization errors of individual channels. With this approach, we can learn activation quantizers that minimize the quantization errors in the majority of channels. Experimental results demonstrate that our proposed method achieves state-of-the-art performance on the image classification task, using AlexNet, ResNet and MobileNetV2 architectures on CIFAR-100 and ImageNet datasets.

\end{abstract}
\section{Introduction}
Deep Convolutional Neural Networks (CNNs) have been playing a crucial role in the recent tremendous successes of a variety of computer vision tasks including image classification, object detection, segmentation, image retrieval, to name a few. However, unfortunately, this benefit comes at the cost of an excessive amount of memory and expensive computational resources, which impedes its application in embedded devices. 
Therefore, how to make CNN lightweight and practical in terms of memory and computation is an important task and has attracted a great amount of effort.

An effective approach is to use low bit-width weights and/or low bit-width activations. This approach not only can reduce the memory footprint, but it also achieves a significant gain in speed. In particular, using binary weights with full-precision (FP) activations can reduce the model size by $32 \times$ and achieve $\sim\hspace{-3pt}2\times$ faster during inference \cite{xnor_net}. When also using low bit-width activations, the speedup is significantly greater; as the most computationally expensive convolutions can be done by bitwise operations \cite{xnor_net}.
Even though there are great improvements achieved by the existing quantization-based methods \cite{hwgq,lossaware,HORQ,xnor_net,tbn,LQNets,dorefa,TTQ}, there are still noticeable accuracy gaps between the quantized CNNs and their FP counterparts, especially in the challenging cases of 1 or 2 bit-width weights and activations. 

Most of the current state-of-the-art network quantization methods \cite{tbn,LQNets,dorefa,TTQ,zhou:2018} learn the FP weights and the quantizers to quantize FP weights. The learning can be sequential or joint learning. In the sequential learning, the models firstly learn FP weights. They then learn the quantizers to quantize FP weights. In the joint learning, the models learn the FP weights and their quantizers simultaneously. 
However, both learning approaches cause mismatch during training, i.e., the gradient descent process is only used to update the FP weights, but not the quantized weights.
To address this problem, we propose a novel method that allows gradients to update the quantized weights with learnable quantization levels directly, i.e., the quantized weights are learned without requiring FP weights. In specific, the quantized weights are learned via learning two auxiliary variables: binary weight encodings and quantization basis vectors.

Another important aspect of training a low bit-width network is the activation quantization. Earlier works use simple, hand-crafted quantizers (e.g., uniform or logarithmic quantization) \cite{DL_limited_num_precision,fix_point_quant,dorefa,INQ,QNN} or pre-computed quantizers fixed during network training \cite{hwgq}. Recently, \cite{LQNets,QIL} proposed to learn the quantizers for activations during training adaptively. 
However, all existing methods consider all activation channels of an activation map 
equally and do not pay attention to the fact that different channels of activations contain different amount of information.
As a result, the learned quantizer could be biased toward a few channels with high variances; while losing information from a large number of lower-variance channels.
To handle this problem, we propose a novel activation quantization method which tries to minimize the information loss in the majority of channels, which cumulatively have noticeable impacts on outputs, via learning channel-wise quantizers (i.e., one quantizer per channel). 
Our main contributions can be summarized as follows:

\begin{itemize}[leftmargin=10pt]
\item 
Our work is the first one to propose a novel method to learn the quantized weights \red{with learnable quantization levels} directly by using gradient descent such that they directly minimize the cost function. Hence, our method could avoid the training mismatch (i.e., the quantized weights are not updated by gradients) of existing methods.
\item To avoid the activation quantizers being biased toward a few high variance activation channels, we propose to learn quantizers that focus  
to minimize quantization errors of the majority of channels. 
This is obtained via learning channel-wise quantizers. 
\item A detailed analysis on the  inference efficiency of the proposed method is provided. The proposed method also achieves the state-of-the-art image classification accuracy when comparing to other low bit-width networks on CIFAR-100 and ImageNet datasets with AlexNet, ResNet and MobileNetV2 architectures. 
\end{itemize}

\section{Related Works}
How to improve inference efficiency of CNN for practical applications has been an active research field. 
Existing approaches can be roughly divided into three main categories: {compact network design},
{parameter reduction,} 
and {network quantization}. In this section, we discuss recent works on the last category, to which our work belongs.

\paragraph{Weight-only Quantization.}
Earlier works have applied the basic form of weight quantization to directly constrains the weight values into the binary space without or with a scaling factor, i.e., $\{-1, 1\}$ in BinaryConnect \cite{BinaryConnect} or $\{-\alpha, \alpha\}$ in Binary-Weight-Networks (BWN) \cite{xnor_net}. 
Ternary Weight Networks (TWN) \cite{TWN} proposed to quantize weights into a ternary form of $\{-\alpha, 0, \alpha\}$. Trained Ternary Quantization (TTQ) \cite{TTQ} generalized TWN by constraining the weights to asymmetric ternary values $\{-\alpha_n, 0, \alpha_p\}$. In comparison to the binary weights, the ternary weights can help to improve performance, while achieving a similar speedup during inference \cite{tbn}. 
\citeauthor{lossaware} proposed loss-aware quantized networks which can further improve the accuracy by considering the loss in learning the binary/ternary weights. \citeauthor{INQ} presented incremental network quantization (INQ) to incrementally convert a pre-trained FP CNN model to a low-precision one.

\paragraph{Weight and Activation Quantization.}
Since quantization of activations can substantially reduce complexity further, by allowing the dot products to be implemented with bit-wise operations (i.e., ${xor}$ and ${popcount}$) \cite{BinaryNet,xnor_net}, this approach attracts more and more attention \cite{xnor_net,dorefa,LQNets}. 
In early works \cite{BinaryNet,xnor_net}, the authors proposed to binarize both weights and activations to $\{-1,1\}$. Although these methods can be highly efficient in reference time, there are considerable accuracy drops compared to FP networks. 
To address this problem, the generalized low bit-width quantization was further studied in \cite{dorefa,log_data_represent}. By allowing to use more bits, this approach can provide a trade-off between accuracy and inference complexity.
A popular choice of  early works is the uniform quantization \cite{dorefa,QNN}.
\citeauthor{log_data_represent} later proposed logarithmic quantization which can help to improve the inference efficiency via bit-shift operations. 
\citeauthor{weighted_entropy} used weighted-entropy to learn quantizers that are more concentrated on the values that are neither too small nor too large.
\citeauthor{hwgq} proposed to pre-compute a single activation quantizer for all activation layers by fitting the probability density function of a half-wave Gaussian distribution. 
As opposed to the fixed, handcrafted quantization schemes of the aforementioned works, LQ-Nets \cite{LQNets} proposed to jointly train a quantized CNN and its associated quantizers.
\citeauthor{QIL} proposed to train quantizers with parameterized intervals, which simultaneously performs both pruning and clipping. \citeauthor{bonn} proposed  Bayesian optimized
1-bit CNNs (BONN) which incorporates the prior distributions of FP kernels and features into the Bayesian framework to construct 1-bit CNNs.
%
Besides, recent works on designing new structures for binary networks achieve  promising results.
Bi-Real Net \cite{bi-real} introduced a new variant of residual structure to preserve the real activations before the $\sign$ function. The shortcut can help to increase the representational capability of the 1-bit convolutional block. 
\citeauthor{groupnet} proposed Group-Net, in which a set of binary convolution branches can effectively replace a FP convolution.


\section{Proposed Method}

\subsection{Learned Quantized Weights}
Apart from existing network quantization works, which train the weight quantizers from the full-precision (FP) network weights either jointly or in advance \cite{dorefa,tbn,LQNets}; in this work, we propose to directly learn the quantized weights. 

Specifically, for a general case of representing a network weight $\bW_q$ of $N_w$ elements by $K_w$-bits, we learn a binary weight encoding $\bS_b^w\in\{-1,+1\}^{N_w\times K_w}$ and a quantization basis vector $\bv^w=[v_1^w,\cdots, v_{K_w}^w]\in\bbR^{K_w}$. Consequently, the network weight can be directly replaced by $\bS_b^w\bv^w$ in the cost function, i.e., $\bW_q=\bS_b^w\bv^w$.
\begin{equation}
    \min_{\Theta}\mathcal{L}(\Theta),\quad \textrm{s.t. } \bS_{bi}^w\in\{-1,+1\}^{N_w\times K_w} ~~\forall i,
\end{equation}
where $\Theta=\{\bS_{bi}^w,\bv_i^w\}_{\forall i}$ is the set of all network parameters and $\mathcal{L}$ is a cost function, e.g., cross-entropy for classification. 

\begin{algorithm}[t]
\setlength{\textfloatsep}{-10pt}
\setlength{\floatsep}{-10pt}
\setlength{\intextsep}{0pt}
\caption{\textbf{Learned Quantized Weights (LQW)}.\newline $\mathcal{L}$ is the  cost function for a mini-batch.}\label{algo:train_quantized_weight}
\SetKwInOut{Requires}{Requires}
\SetKwInOut{Output}{Output}
\SetKwInOut{Parameters}{Parameters}
\Requires{Learning rate $\eta$ and learning-rate scaling factors $\{\gamma_{\bv}, \gamma_{\bs}\}$}
\Parameters{Encoding matrix $\bS^w$\hspace{-2pt}, basis vector $\bv^w$}
\textbf{Forward propagation:} \\
$\quad\bS_b^w=\sign(\bS^w)$\\
$\quad\bW_q=\bS_b^w\bv^w$\\
\textbf{Backward propagation and Parameter update:}\\
$\quad$Given $\frac{\partial\mathcal{L}}{\partial\bW_q}$ obtained using the chain rule of gradient descent, compute $\frac{\partial\mathcal{L}}{\partial\bS_b^w}$ and $\frac{\partial\mathcal{L}}{\partial\bv^w}$\\
$\quad\bv^w\gets\bv^w - \gamma_\bv\eta\frac{\partial\mathcal{L}}{\partial\bv^w}$\\
$\quad\bS^w\gets\textrm{clip}\left(\bS^w - \gamma_\bs\eta\frac{\partial\mathcal{L}}{\partial\bS_b^w}, -1, 1\right)$ ~~~~~~~~~~\tcp{STE}
\end{algorithm}

To handle the binary constraint on the weight encoding $\bS_b^w$, inspired by \cite{BinaryConnect}, we 
introduce a full-precision encoding matrix $\bS^w$ (i.e., $\bS^w\in\bbR^{N_w\times K_w}$) for parameter updates, and obtain $\bS_b^w$ using $\sign$ function (i.e., $\bS_b^w=\sign(\bS^w)$) only during forward and backward propagations.
Furthermore, in order to update $\bS^w$ using gradient descent, we adopt the straight-through estimator (STE) $\frac{\partial\sign(z)}{\partial z}=1_{|z|\le 1}$ \cite{STE}, i.e., $\frac{\partial \mathcal{L}}{\partial \bS_b^w}=\frac{\partial \mathcal{L}}{\partial \bS^w}$ for $\bS^w\in[-1, 1]$, to approximate the gradients propagating through the $\sign$ function. 
Regarding the quantization basis vectors $\bv^w$, as they have no constraint, they can be updated normally using the standard gradient descent process.
The proposed method allows the quantized weights \red{with learnable quantization levels} to be updated (via the weight encodings and the quantization basis vectors) such that they directly minimize the final cost function through the gradient descent process. Additionally, we want to emphasize that other works (e.g., LQ-Nets, TBN) achieve quantized weights from FP weights during a joint training process. More specifically, their FP weights are updated via back-propagation, while the quantized weights ($\bW_q$) are updated to minimize the $\ell_2$-loss with the learning FP weights (i.e., $\arg\min_{\bW_q}\|\bW_q-\bW\|^2$). This process potentially causes training mismatch, as the gradients cannot directly affect the quantized weights.   

Algorithm \ref{algo:train_quantized_weight}, dubbed as Learned Quantized Weights (LQW), presents the procedure of our proposed method to train a quantized network weight. 
Note that, to maximize the flexibility of the low bit-width CNN, while ensuring its compatibility with bitwise operations, we learn one quantization basis vector and weight encoding per filter.


\begin{figure*}[ht]
\centering
\begin{subfigure}[b]{0.23\textwidth}
\includegraphics[width=\textwidth]{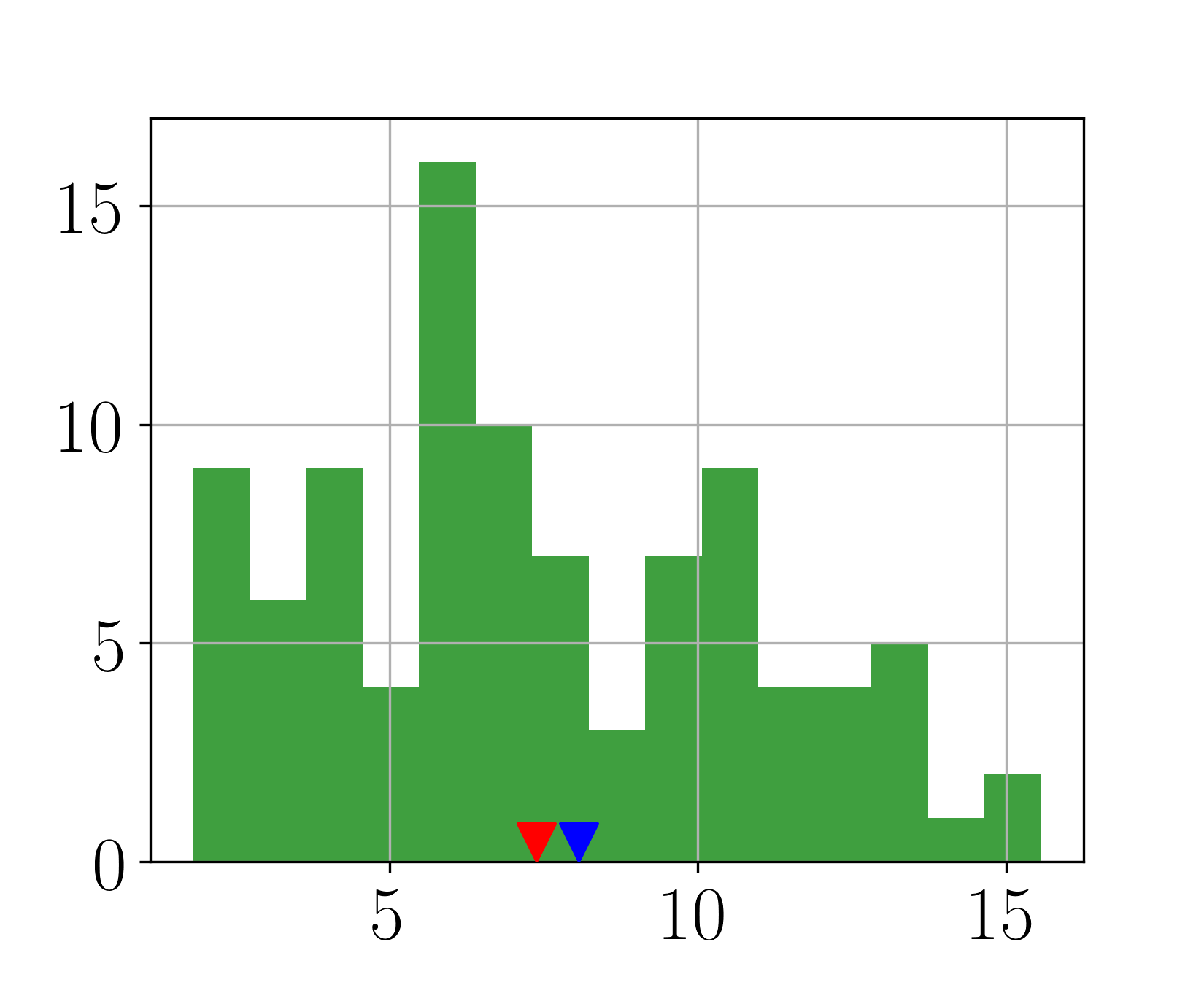}
\caption{$2$-nd quant. level}
\label{fig:2nd_quant}
\end{subfigure}
~~
\begin{subfigure}[b]{0.23\textwidth}
\includegraphics[width=\textwidth]{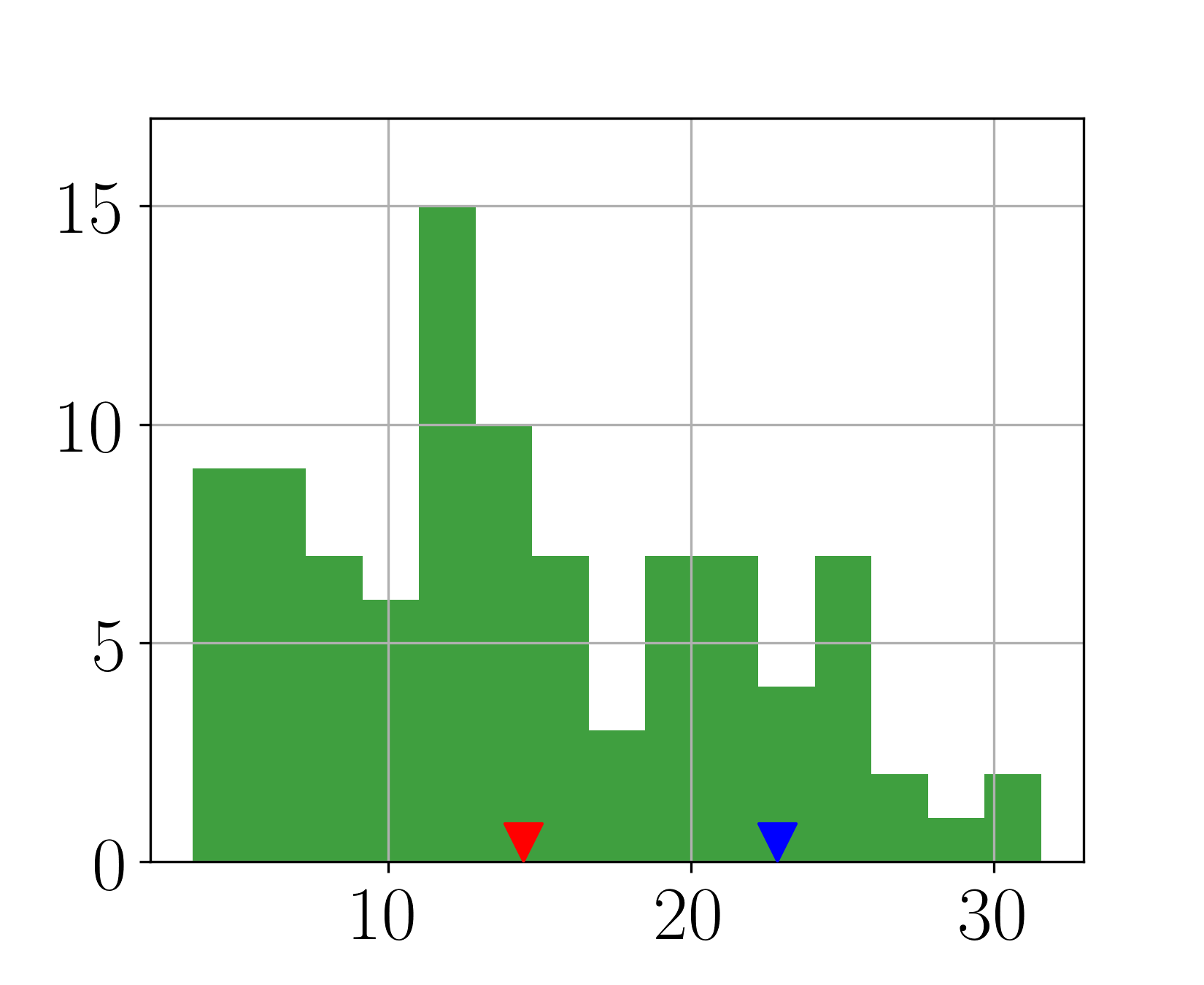}
\caption{$3$-rd quant. level}
\label{fig:3rd_quant}
\end{subfigure}
~~
\begin{subfigure}[b]{0.23\textwidth}
\includegraphics[width=\textwidth]{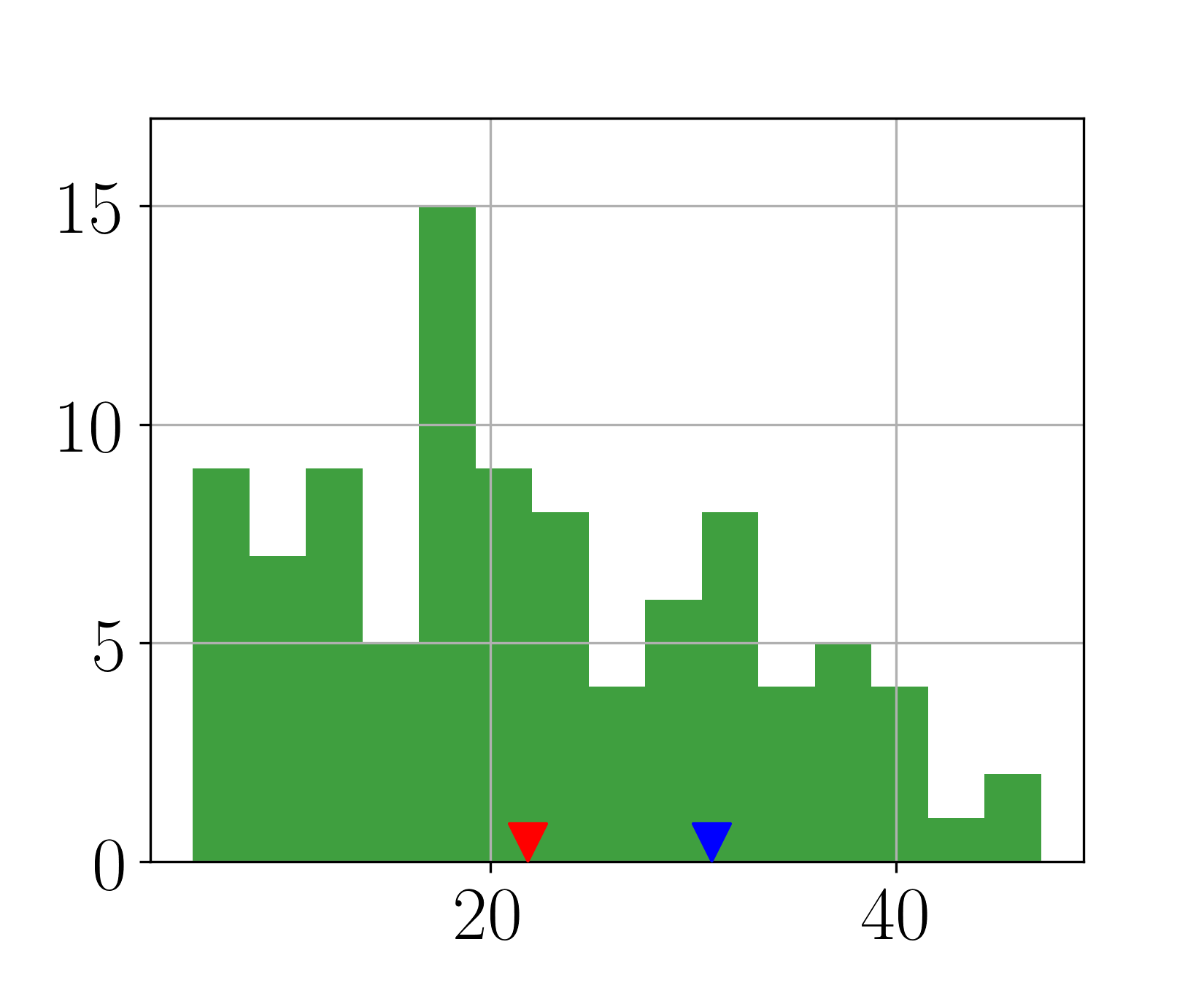}
\caption{$4$-th quant. level}
\label{fig:4th_quant}
\end{subfigure}
~~
\begin{subfigure}[b]{0.23\textwidth}
\includegraphics[width=\textwidth]{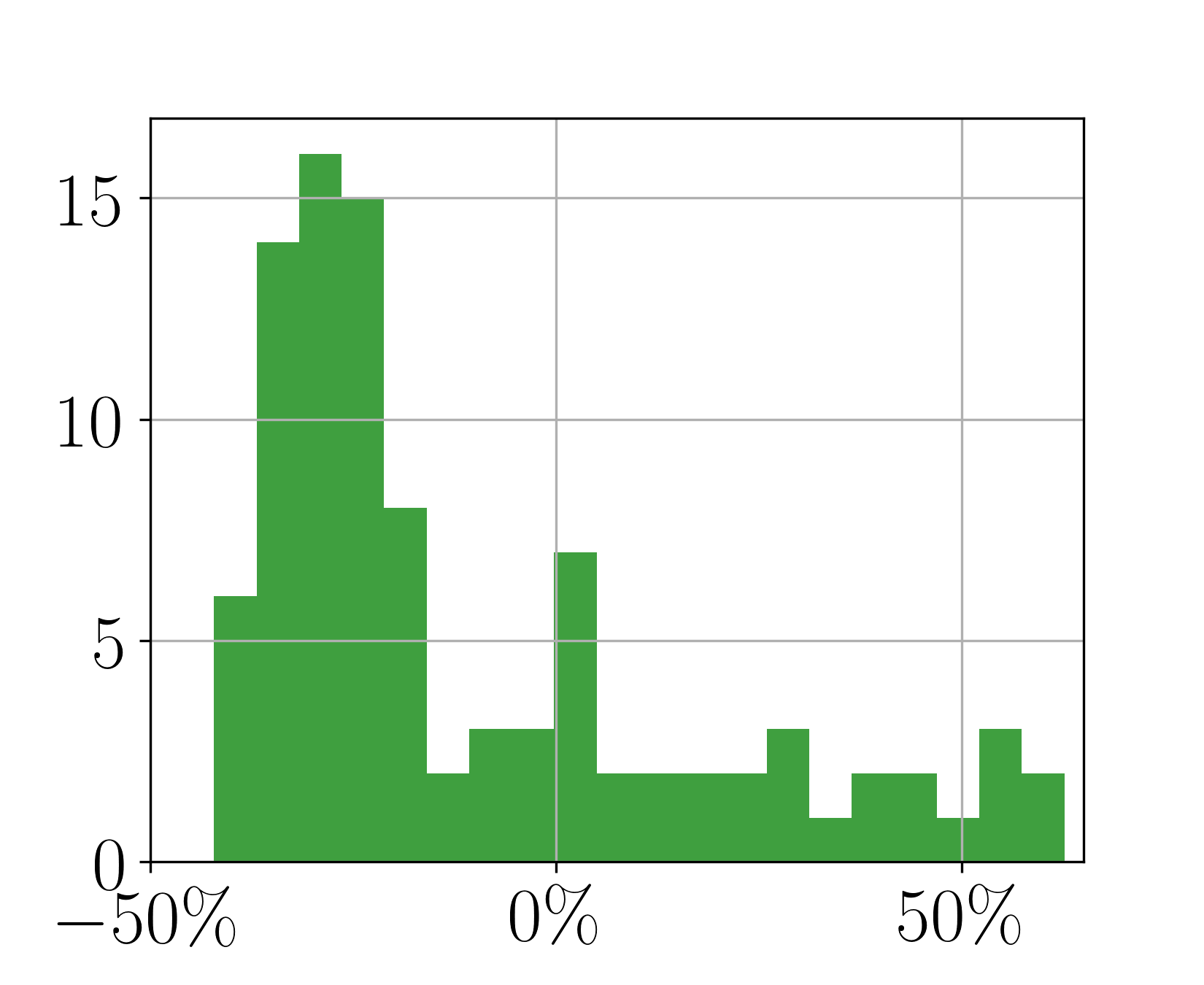}
\caption{MSE relative changes}
\label{fig:mse_change}
\end{subfigure}
\vspace{-3pt}
\caption[Fig. \ref{fig:2nd_quant}, \ref{fig:3rd_quant}, and \ref{fig:4th_quant} are respectively the distributions of the optimal $2$-nd, $3$-rd, and $4$-th channel-wise quantization levels of the first activation layer of AlexNet with $K_a=2$.
{\small\color{red}\ding{116}} indicates the proposed channel-wise averaged quantization levels - Mean Square Error (MSE) = $7.376$ (using 100 random ImageNet images); and {\small\color{blue}\ding{116}} indicates the quantization levels when considering the whole activation as the input for a scalar quantizer (called as baseline)
- MSE = $6.664$.  
Fig. \ref{fig:mse_change} shows the histogram of the relative changes (\%) in channel-wise MSE of our method compared to the baseline.
We can observe that, in the baseline approach, the quantization levels tend to be biased toward large values. Although our proposed method results in higher quantization error overall (i.e., $7.376>6.664$), the majority of channels (68 out of 96 channels in Fig. \ref{fig:mse_change}) achieve smaller quantization errors (the relative changes $<0$) compared to the baseline.]
{
Fig. \ref{fig:2nd_quant}, \ref{fig:3rd_quant}, and \ref{fig:4th_quant} are respectively the distributions of the optimal $2$-nd, $3$-rd, and $4$-th channel-wise quantization levels of the first activation layer of AlexNet with $K_a=2$.
{\small\color{red}\ding{116}} indicates the proposed channel-wise averaged quantization levels - Mean Square Error (MSE) = $7.376$ (using 100 random ImageNet images); and {\small\color{blue}\ding{116}} indicates the quantization levels when considering the whole activation as the input for a scalar quantizer (e.g., \protect\cite{LQNets}) (called as baseline)
- MSE = $6.664$.  
Fig. \ref{fig:mse_change} shows the histogram of the relative changes (\%) in channel-wise MSE of our method compared to the baseline.
We can observe that, in the baseline approach, the quantization levels tend to be biased toward large values. Although our proposed method results in higher quantization error overall (i.e., $7.376>6.664$), the majority of channels (68 out of 96 channels in Fig. \ref{fig:mse_change}) achieve smaller quantization errors (the relative changes $<0$) compared to the baseline.
}
\label{fig:CAQ}
\vspace{-8pt}
\end{figure*}

\begin{algorithm}[!t]
\caption{\textbf{Channel-wise Averaged Quantizer (CAQ)}. $\mathcal{L}$ is a cost function for a mini-batch.}\label{algo:train_act_quantizer} 
\SetKwInOut{Requires}{Requires}
\SetKwInOut{Parameters}{Parameters}
\SetKwInOut{Input}{Inputs}
\SetKwInOut{Output}{Outputs}
\Input{A mini-batch activation $\bA$ with $C$ channels, Moving-average factor $\mu$, and number of iteration $T$.}
\Output{The quantized activation  $\bA_q$.}
\Parameters{A set of quant. basis vectors $\{\bv^{(j)}\}_{j=1}^C$. }
\If{ in the training stage, }{
\textbf{Forward propagation:} \\
\For
{$j=1$ \KwTo $C$} {
    Set $\bv_0^{(j)} = \bv^{(j)}$ \\
    \For{$t=1$ \KwTo $T$}{
    Given $\bv_{t-1}^{(j)}$, compute $\bS_{t}^{(j)}$ by looking up the index of the nearest quant. level;\\
    Given $\bS_{t}^{(j)}$, compute $\bv_{t}^{(j)}$ by using Eq. \ref{eq:update_v}.
    } 
    $\bv^{(j)}\gets (1-\mu)\bv_{T}^{(j)}+\mu\bv^{(j)}$\\
}
$\bv^a\hspace{-1pt}=\hspace{-1pt}\frac{1}{C}\hspace{-1pt}\sum_{j=1}^C\hspace{-2pt}\bv^{(j)}$\\
Compute $\bA_q$ by looking up the nearest quantization level 
 $\in \bv^a$ for each element.\\
\textbf{Backward propagation:}\\
Given $\frac{\partial\mathcal{L}}{\partial\bA_q}$ obtained using back-propagation;\\
$\frac{\partial \mathcal{L}}{\partial \bA}= \frac{\partial \mathcal{L}}{\partial \bA_q}\quad$ ~~~~~~~~~~~~~~~~~~~\tcp{STE}
}
\Else(\tcp*[h]{inference stage}){
$\bv^a=\frac{1}{C}\sum_{j=1}^C\bv^{(j)}$ 
\\
Compute $\bA_q$ by looking up the nearest quantization level $\in \bv^a$ for each element.\\
}
\end{algorithm}

\subsection{Activation Channel-Wise Averaged Quantizer}
\label{sec:act_quant}
%

\paragraph{Scalar Quantizer.}
Given a full-precision ReLU activation $\bA\hspace{-0.5pt}\in\hspace{-0.5pt}\bbR_{\ge 0}^{B\times C\times W\times H}$ (i.e., the set of non-negative real values), our goal is to represent $\bA$ using $K_a$-bits with minimized information loss. 
We consider $\bA$ as a vector $\ba$ of $N_a$-dimension ($N_a\hspace{-1pt}=\hspace{-1pt}BCWH$). 
We aim to find an optimal quantization basis vector $\bv^a\hspace{-1pt}=\hspace{-1pt}[v_1^a,\cdots,v_{K_a}^a]\hspace{-1pt}\in\hspace{-1pt}\bbR^{K_a}$ and an optimal binary encoding $\bS^a\in\{0,1\}^{N_a\times K_a}$ that minimize the following quantization error:
\begin{equation}
\label{eq:act_quantizer}
    \arg\min_{\bv^a,\bS^a}\|\ba-\bS^a\bv^a\|_2^2,\quad\textrm{s.t. }\bS^a\in\{0,1\}^{N_a\times K_a}.
\end{equation}
This problem can be solved by alternatively updating $\bS^a$ and $\bv^a$:
\textbf{\textit{(i)}} Given $\bv^a$ fixed, the optimal $\bS^a$ can be found by looking up the index of the nearest quantization level. 
Note that, for a $K_a$-bit quantization, there are $2^{K_a}$ quantization levels $\bq^a=[q_i^a,\cdots,q_{2^{(K_a)}}^a]\in\bbR^{2^{(K_a)}}$ and 
    $q_i^a = \left\langle\text{Dec2Bin}_{K_a}(i), \bv^a\right\rangle,$
where $\langle\cdot\rangle$ is a dot product and $\text{Dec2Bin}_{K_a}(i)$ is the function convert the decimal value $i$ ($0\le i<2^{K_a})$ to the equivalent $K_a$-bit binary vector, e.g., $\text{Dec2Bin}_{3}(6)=[0, 1, 1]$.
%
\textbf{\textit{(ii)}} Given $\bS^a$ fixed, we have the closed-form solution for $\bv^a$ as follows: 
\begin{equation}
\label{eq:update_v}
\bv^a = ({\bS^a}^\top\bS^a)^{-1}{\bS^a}^\top\ba.   
\end{equation}
\paragraph{Channel-Wise Averaged Quantization.}
The Batch Normalization layer \cite{batch_norm} includes two steps: {{(i)}} normalizing and {{(ii)}} shifting and scaling. While the normalizing step tries to make each channel of $\bA$ to have zero-mean and unit-variance, the scaling and shifting step makes the variances of channels different. The variances can be very large or very small, depending on the corresponding scaling factor of the $i$-th channel.
Hence, when a single scalar quantizer is used for $\bA$ as in previous works (e.g., LQ-Nets \cite{LQNets}) (called as baseline), few channels which have high variances can significantly affect the quantization basis values. Consequently, the resulting quantization basis vector potentially causes significant information loss in many channels with low variances, which cumulatively have noticeable impacts on outputs. 
Additionally, slightly less focus on large activation values might help to reduce overfitting.
To address this problem, first, we propose to learn a scalar quantizer for each channel of activation $\bA$. Then, the quantization basis values for the layer-wise quantizer can be obtained by taking the average of all quantization basis values of channel-wise quantizers, i.e., $v_i^a=\frac{1}{C}\sum_{j=1}^Cv_i^{a(j)},$  where $v_i^{a(j)}$ is the $i$-th quantization basis value of $j$-th channel. With this proposed approach, the layer-wise quantization levels minimize the quantization errors on the majority of channels. 
Figure \ref{fig:CAQ} provides an example to illustrate our proposed method.
Besides, noticeably, only a mini-batch of input is available at each forward/backward step, we necessarily apply exponential moving average to update the channel-wise quantization level vectors to ensure stability.
Algorithm \ref{algo:train_act_quantizer}, dubbed as Channel-wise Averaged Quantizer (CAQ), summarizes the procedure for training and inference of our proposed quantizer for activations. Note that during inference, only the channel-wise averaged quantizer is required. Hence, our method has the same time complexity as the baseline does.

\vspace{-3.5pt}
\section{Experiments}
\vspace{-1pt}
To evaluate the performance of the proposed methods, we conduct experiments on CIFAR-100 \cite{cifar10} and ImageNet (ILSVRC2012) \cite{imagenet} datasets using the two representative network architectures, AlexNet \cite{alexnet}, ResNet \cite{resnet}, and \red{MobileNetV2 \cite{MobileNetV2}}. We adopt the standard data splits for both datasets, i.e., 50K training and 10K test images for CIFAR-100, about 1.2 million training and 50K test images for ImageNet.
Additionally, we use a variant of AlexNet architecture by adding Batch Normalization layers after each convolutional layer and Fully-Connected layer and removing the Local Response layers as \red{commonly used in recent works \cite{dorefa,guidedQuantize}.} 

\vspace{-3pt}
\subsection{Implementation Details}
We follow the common settings in previous works \cite{dorefa,LQNets}: We first resize the shorter side of the images to $256$. During training, we randomly crop  $227\times 227$/$224\times 224$ patches for AlexNet/ResNet and MobileNetV2 from training images or their horizontal flips (called as basic augmentation). Following \citeauthor{LQNets}, we adopt the basic augmentation in most of our experiments, except for the cases of ResNet on ImageNet dataset with $K_w/K_a=2/2,3/3$, where the augmentation strategy of the ResNet Torch implementation\footnote{\url{https://github.com/facebook/fb.resnet.torch}} is adopted. We report top-1 and top-5 classification accuracy using single-center crops of test images. 

In all experiments, we use Nesterov SGD with a momentum of $0.9$ and  mini-batch size of $256$. The starting learning rate is set at $0.02 / 0.1/0.02$ for AlexNet/ResNet\red{/MobileNetV2} respectively. We adopt the polynomial learning rate scheduler with the annealing power of 2 and the final learning rate of $10^{-6}$ after $120$ epochs. Regarding the weight decay, for FP weights, we set weight decay as $5 \times  10^{-4}$/$5 \times  10^{-5}$/$10^{-5}$ for AlexNet/ResNet\red{/MobileNetV2}. For the quantized weights, the weight decay has a stronger effect; hence, we cross-validate within $\{10^{-6},5 \times  10^{-6},$ $ 10^{-5}, 5 \times  10^{-5}\}$ using 1\% of the training set as the validation set.
We scale the learning rate of the binary weight encodings as suggested by \citeauthor{BinaryConnect}. 
We also scale the learning rate of quantization basis vectors by a factor of $\gamma_\bv=\frac{1}{50}$ to avoid unstable training. The moving-average factor $\mu$ and the number of iteration $T$ in CAQ algorithm are fixed as $0.9$ and $1$, respectively.
Following previous works \cite{xnor_net,dorefa,LQNets}, we do not quantize the first and last layers. Since the speedup, benefited by bitwise operations, is low due to their small input-channel number or filter size \cite{dorefa,xnor_net}.   
All networks using our proposed method are trained from scratch.  

\begin{figure*}[]
\centering

\begin{subfigure}[b]{0.23\textwidth}
\includegraphics[width=\textwidth]{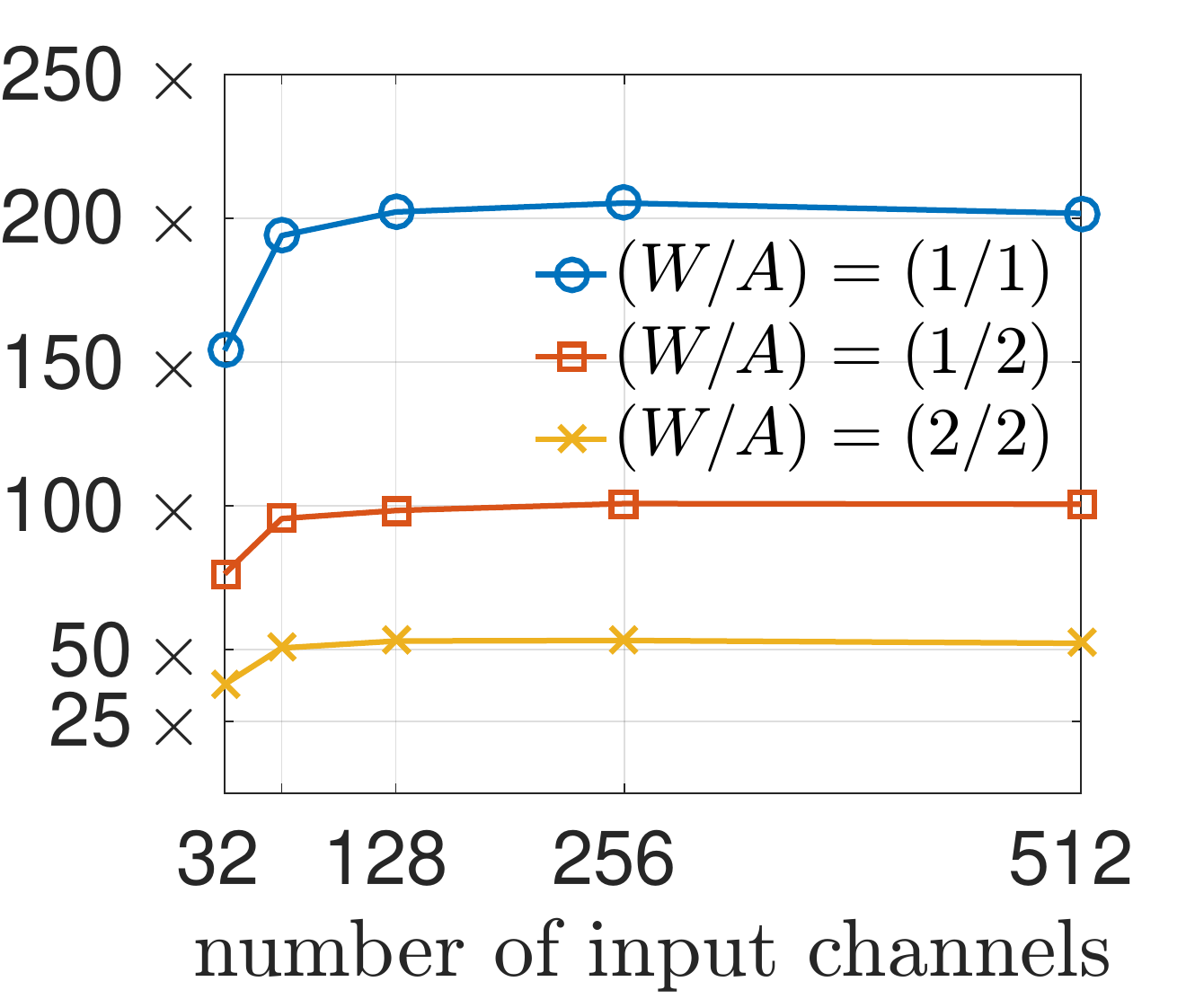}
\caption{1 thread w/o SIMD}
\label{fig:speedup_cpp}
\end{subfigure}
$\qquad$
\begin{subfigure}[b]{0.23\textwidth}
\includegraphics[width=\textwidth]{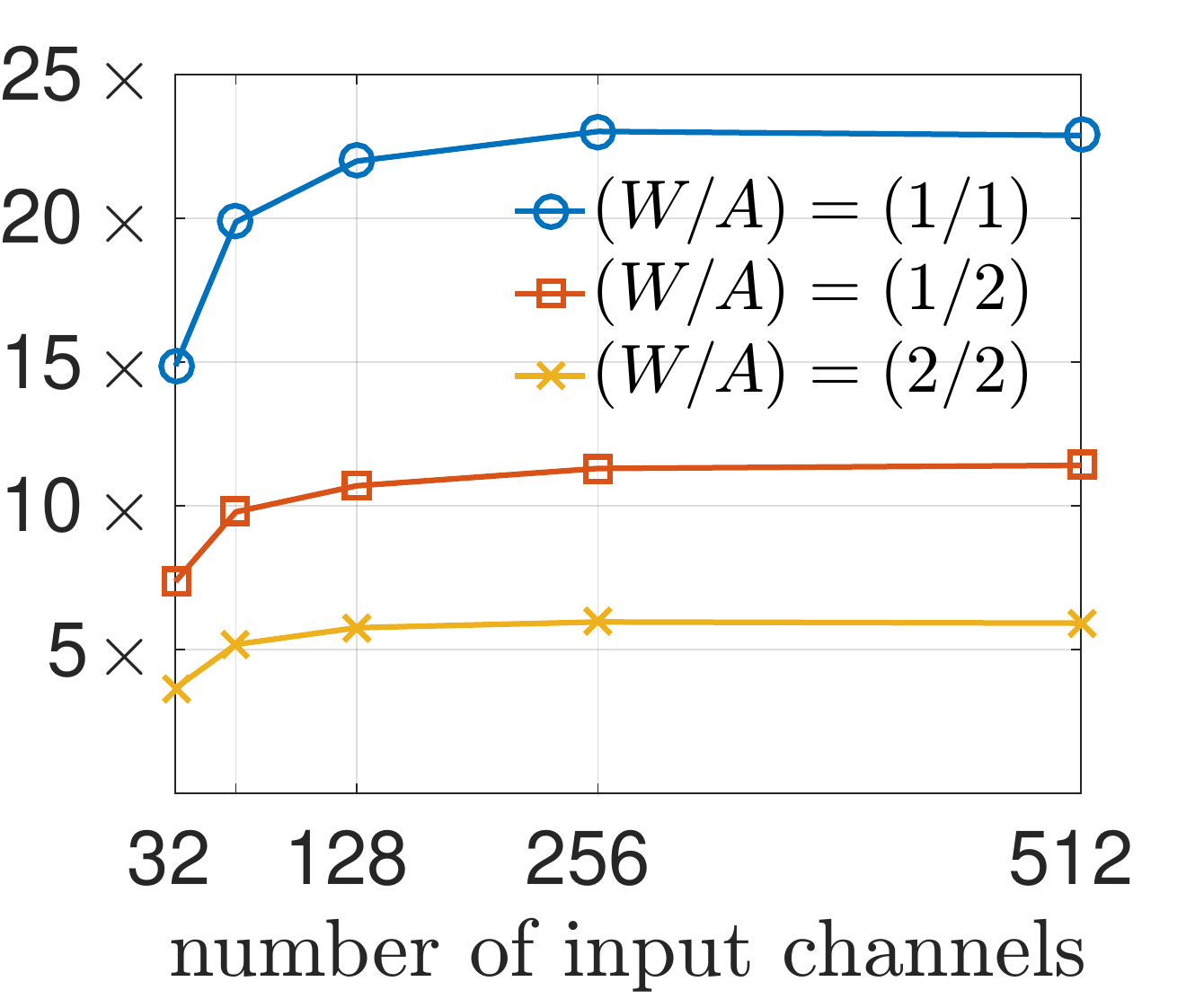}
\caption{1 thread with SIMD}
\label{fig:speedup_simd}
\end{subfigure}
$\qquad$
\begin{subfigure}[b]{0.23\textwidth}
\includegraphics[width=\textwidth]{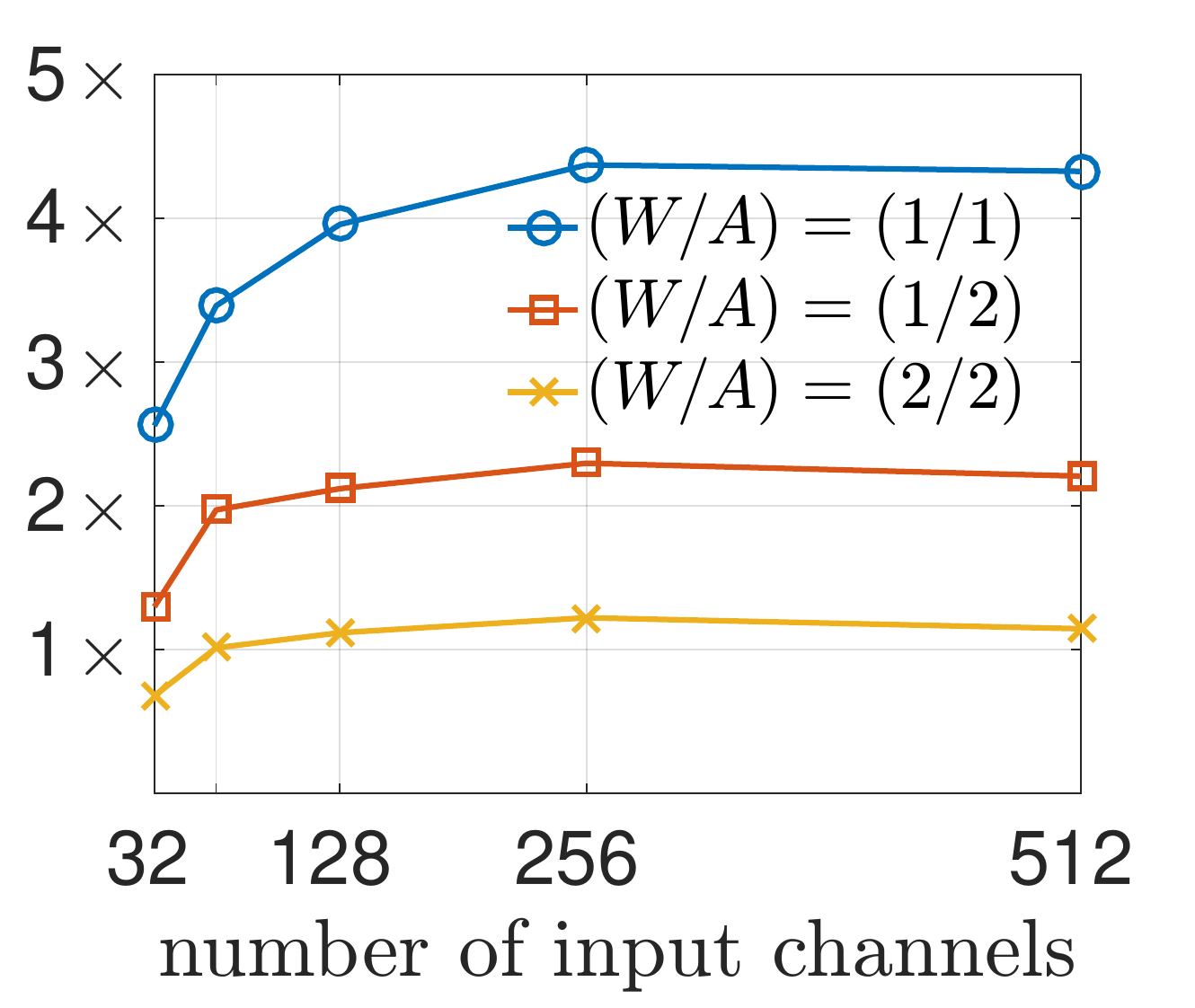}
\caption{CBLAS (8 threads)}
\label{fig:speedup_blas}
\end{subfigure}
\vspace{-5pt}
\caption{The speedup of one convolution by varying input channel size at different bit-widths.}
\label{fig:speedup}
\vspace{-9pt}
\end{figure*}

\subsection{Inference Efficiency Analysis}
\label{sec:efficicency_analysis}
We first show that the inner products between our quantized weight and activation vectors can be efficiently computed by bitwise operations. Let a weight vector $\bW \in \bbR^N$ be encoded by $\bS^{w}=\{\bs_1^{w},\cdots,\bs_{K_w}^{w}\} \in \{-1, 1\}^{N\times K_w}$ and the learned basis vector $\bv^w=\{v_1^w,\cdots, v_{K_w}^w\} \in \bbR^{K_w}$.
Similarly, we encode an activation vector $\bA \in \bbR^N$ by $\bS^{a}=\{\bs_1^{a},\cdots,\bs_{K_a}^{a}\}\in \{0, 1\}^{N\times K_a}$ and $\bv^a=\{v^a_1,\cdots,v^a_{{K_a}}\}$ $\in \bbR^{{K_a}}$.
To utilize the bitwise inner products \cite{dorefa}; first we need to convert $\bS^{a}\in \{0, 1\}^{N\times K_a}$ to $\hat{\bS}^{a}\in\{-1,1\}^{N\times K_a}$ by $\bS_i^{a} = 0.5\hat{\bS}_i^{a} +0.5$.
Then, with some manipulation, we can derive that:
\begin{equation}
\label{eq:bitwise_dot_product}
\hspace{-5pt}\mathcal{Q}(\bW,\bv^w)^\top\mathcal{Q}(\bA, \bv^a)=Q+\hspace{-2pt} \sum_{i=1}^{K_w}\sum_{j=1}^{{K_a}}\frac{v_i^wv_j^a}{2}\left(\bs_i^{w}\odot\hat{\bs}_j^{a}\right),
\end{equation}
where $Q = \sum_{i=1}^{K_w}\sum_{j=1}^{{K_a}}\frac{v_i^wv_j^a}{2}\left(\bs_i^{w}\odot \boldsymbol{1}\right)$, $\boldsymbol{1}$ is a $K_a$-dimension vector with all elements are 1, and $\odot$ denotes the inner product with bitwise operations, \textit{xnor} and \textit{popcnt}. Noting that during inference, $\bS^w$, $\bv^w$, and $\bv^a$ are fixed; so $Q$ is a scalar constant and can be pre-computed. Similar to \cite{dorefa,LQNets}, the computation complexity of Eq. \ref{eq:bitwise_dot_product} w.r.t. $K_w,K_a$ is $\mathcal{O}(K_w{K_a})$, i.e., directly linear-proportional to the bitwidths of weights and activations. 

Considering a matrix multiplication (MatMul) $\widetilde{\bC}=\widetilde{\bW}\widetilde{\bA}$, where $\widetilde{\bW}\in \bbR^{n\times p}$, $\widetilde{\bA}\in \bbR^{p\times m}$ and  $\widetilde{\bC}\in \bbR^{n\times m}$. 
To calculate $\widetilde{\bC}$ with FP $\widetilde{\bW}, \widetilde{\bA}$, there are $n m q$ multiply-accumulate operations (MACs) required.
Following \citeauthor{tbn}, the theoretical speedup ratio for the bitwise MatMul of our method, which requires $K_w{K_a} mn$ MACs and $2(K_w{K_a})mn$ binary operations, is given as follows:
\begin{equation}
\label{eq:speedup}
S = \frac{\gamma q}{\gamma K_w{K_a}+2(K_w{K_a})\lceil\frac{q}{L}\rceil},
\end{equation}
where $L$ is the number of bits binary operation in one clock cycle, and  $\gamma = 1.91$ is ratio between the speed of performing a $L$-bit binary operation and a MAC \cite{tbn}.

However, we found that Eq. \ref{eq:speedup} may not accurately reflect the actual speedup ratio of MatMul using bitwise operations in comparison with using FP.
First, Eq. \ref{eq:speedup} over-simplifies the bitwise MatMul by only including \textit{xnor} and \textit{popcount} operations. In fact, additional operations are required, including \textit{sum} to accumulate the \textit{popcount} results, operations to convert from integer (\textit{popcount} results) to FP numbers to multiply with quantization basis values. 
Second, Eq. \ref{eq:speedup} under-utilizes the CPU capability with FP computation. The modern CPU can perform 4 MACs in a single clock by using Single Instruction Multiple Data (SIMD) instructions. Multi-threading can also be utilized to further speed up computation by dividing the workload into multiple separate cores.   
Third, data transferring operations (e.g., loading, storing) is not considered. 
Using quantized weights and activations or SIMD can help to reduce the number of loading/storing operations. 
Finally, the additional time for quantizing activations should be considered in computing the speedup ratio.

We provide a CPU implementation of the bitwise MatMul (Eq. \ref{eq:bitwise_dot_product}) to measure its actual CPU speedup in comparison with FP MatMul. We fix $n=256$ (i.e., the number of output channels $C_o$), $m=14 \times  14 \times  100$ (i.e., an 100-sample mini-batch of $14 \times  14$ size), and $3 \times 3$ kernel. Noting that $p=3\times 3\times C_i$, where $C_i$ is the number of input channels.
The experiment is conducted on a quad-core CPU (i7-6700HQ@2.60GHz) with $L=64$. In Figure \ref{fig:speedup}, we present the empirical speedup ratio of the bitwise MatMul in compared with the FP MatMul (with different implementations of FP MatMul) as the number of input channels $C_i$ varies.

Firstly, from Eq. \ref{eq:speedup}, we can compute the theoretical speedup for $K_w/K_a=1/1$ is $\sim\hspace{-2pt}60\times$. However, our CPU implementation of bitwise MatMul, with all overheads including quantization and memory allocation, can achieve up to $\sim\hspace{-2pt}200\times$ speedup (Figure \ref{fig:speedup_cpp}) (when SIMD is not utilized for FP MatMul). 
When SIMD is utilized for FP MatMul, bitwise MatMul can achieve about $\sim\hspace{-2pt}23\times$ speedup for $K_w/K_a=1/1$ (Figure \ref{fig:speedup_simd}).
Additionally, when multi-threading (8 threads) is also considered for both FP (CBLAS library)
and bitwise MatMul, bitwise MatMul can gain $\sim\hspace{-2pt} 4.3\times$ speedup for $K_w/K_a=1/1$. While for $K_w/K_a=2/2$, we obverse only a small speedup for bitwise MatMul (for $C_i>64$) (Figure \ref{fig:speedup_blas}).
Furthermore, the empirical speedup ratios confirm that the complexity of our bitwise MatMul is proportional to the bit-widths of weights and activations.

\red{Additionally, we presented in Table \ref{tb:reference_time} the inference time per image using bitwise convolution for the quantized AlexNet (for both weight and activation) models. We also include the inference time using FP convolution operation for comparison. We can see that when using AlexNet with ${K_w}/{K_a}=1/2$, our proposed method with bitwise convolution can speed up $\times$1.7 and $\times$6.8 in comparison with FP convolution with and without using SIMD, respectively.}

\begin{table}[]
\setlength{\tabcolsep}{8pt}
\centering
\begin{tabular}{|c|c|c|c|}
\hline
\multicolumn{2}{|c|}{FP} & \multicolumn{2}{c|}{Bitwise} \\ \hline
W/o SIMD & With SIMD & 2 / 2 & 1 / 2 \\ \hline
3254$\pm$28 & 809$\pm$19 & 523$\pm$5 & 475$\pm$8 \\\hline
\end{tabular}
\vspace{-4pt}
\caption{The average $\pm$ std of inference time (ms) for 1 image using AlexNet trained on CIFAR-100 dataset under FP and bit-wise conv. implementations. 
The experiments was conducted on a CPU(i7-6700HQ@2.60GHz) with single-thread using 10 random images.
}
\label{tb:reference_time}
\vspace{-7pt}
\end{table}

\begin{table}[t]
\setlength{\tabcolsep}{8pt}
\centering
\begin{tabular}{|c|c|c|}
\hline
Weight & Activation & Top-1  \\\hline
Baseline & Baseline & 69.2 \\
Baseline & \textbf{CAQ} & 69.6 \\
\textbf{LQW} & Baseline  & 69.7 \\
\textbf{LQW} & \textbf{CAQ} & 69.9 \\\hline
\end{tabular}
\vspace{-5pt}
\caption[Top-1 accuracy  (\%) of AlexNet on CIFAR-100 dataset using different methods (LQW, CAQ, and baseline) with $K_w/K_a=2/2$.]
{Top-1 accuracy  (\%) of AlexNet on CIFAR-100 dataset using different methods (LQW, CAQ, and baseline \cite{LQNets}) with $K_w/K_a=2/2$.}
\label{tb:ablation_study}
\vspace{-5pt}
\end{table}

\begin{table}[t]
\centering
\setlength{\tabcolsep}{6pt}
\begin{tabular}{|c|c|c|c|}
\hline
{Bit-widths} & \multicolumn{3}{c|}{Top-1 Accuracy (\% - mean$\pm$std)} \\ \cline{2-4}
{($K^w$/$K^a$)}  &  $T=1$ & $T=2$ & $T=3$ \\ \hline
1 / 2 & 69.42$\pm$0.14 & 69.59$\pm$0.16 & 69.55$\pm$0.12  \\
2 / 2 & 69.88$\pm$0.11 & 70.03$\pm$0.13 & 69.82$\pm$0.09 \\
\hline
\end{tabular}
\vspace{-5pt}
\caption{Accuracy w.r.t. numbers of CAQ iteration $T$}
\label{tab:num_iter}
\vspace{-8pt}
\end{table}

\begin{table*}[t]
\centering
\setlength{\tabcolsep}{7.5pt}
\begin{threeparttable}
\begin{tabular}{|l|c|c|c|c|c|c|c|}
\hline
\multirow{2}{*}{Methods} & {Bit-widths}  & \multicolumn{2}{c|}{AlexNet}  & \multicolumn{2}{c|}{ResNet-18} & \multicolumn{2}{c|}{ \red{MobileNetV2}}  \\ \cline{3-8}
 &  {$({K_w}/{K_a})$} & Top-1 & Top-5 & Top-1 & Top-5 & \red{Top-1} & \red{Top-5} \\ \hline
\multicolumn{2}{|l|}{Full-precision (FP)} & 71.2 & 91.3 & 74.0 & 92.7 & \red{72.9} & \red{92.1} \\ \hline

LQ-Nets  & 3 / 3 & 70.9 & 91.3 & 72.9 & 92.2 & - & -  \\
\textbf{LQW + CAQ} & 3 / 3  & \textbf{71.3} & \textbf{91.3}  & \textbf{73.0} & \textbf{92.3} & \red{\textbf{71.5}} & \red{\textbf{91.6}} \\ \hline
 
PQ+TS+Guided & 2 / 2 & 64.6 & 87.8 & - & - & - & - \\
DoReFa-Net  & 2 / 2 & 63.4 & 87.5 & 65.1 & 88.0 & \red{56.8} & \red{85.2} \\ 
LQ-Nets & 2 / 2 & 69.2 & 91.2 & 70.8 & 91.3 & - & - \\
\textbf{LQW + CAQ} & 2 / 2  & \textbf{69.9} & \textbf{91.3} & \textbf{72.1} & \textbf{92.3} & \red{\textbf{65.5}} & \red{\textbf{89.6}} \\ \hline
 
LQ-Nets & 1 / 2 & 68.7 & 90.5 & 70.4 & 91.2 & - & - \\
\textbf{LQW + CAQ}  & 1 / 2  & \textbf{69.3} & \textbf{91.2} & \textbf{72.1} & \textbf{91.6} & \red{\textbf{63.0}} & \red{\textbf{88.27}} \\ \hline
 
\end{tabular}
\begin{tablenotes}
\footnotesize
\item ``-'' indicates that the results are not provided.
\end{tablenotes}
\end{threeparttable}
\vspace{-6pt}
\caption{Top-1/Top-5 accuracy (\%) of AlexNet, ResNet-18, \red{and MobileNetV2} on CIFAR-100 dataset under different bit-widths. }
\label{tb:compared_cifar100}
\vspace{-4pt}
\end{table*}

\begin{figure}[t]
\centering
\begin{subfigure}[b]{0.235\textwidth}
\includegraphics[width=\textwidth]{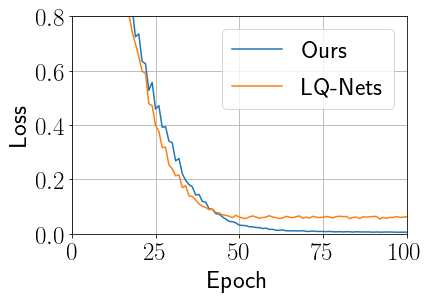}
\caption{Cross-entropy loss}
\label{fig:cross_entropy_loss}
\end{subfigure}
\begin{subfigure}[b]{0.235\textwidth}
\includegraphics[width=\textwidth]{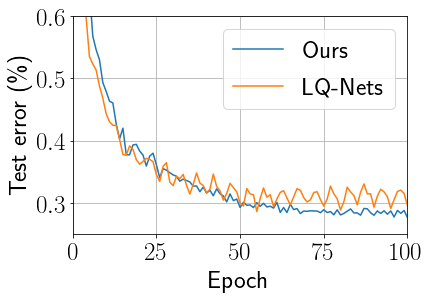}
\caption{Test error}
\label{fig:test_error}
\end{subfigure}
\vspace{-5pt}
\caption{The cross-entropy loss (a) and test error (b) ResNet-18 on CIFAR-100 dataset with $K_w/K_a=1/2$.}
\label{fig:classification_error}
\vspace{-10pt}
\end{figure}


\begin{table*}[ht]
\centering
\setlength{\tabcolsep}{8pt}
\begin{threeparttable}
\begin{tabular}{|l|c|c|c|c|c|c|c|}
\hline
\multirow{2}{*}{Methods} &  {Bit-widths}  & \multicolumn{2}{c|}{AlexNet}  &  \multicolumn{2}{c|}{ResNet-18} & \multicolumn{2}{c|}{ResNet-34}  \\ \cline{3-8}
 & {($K_w$/$K_a$)} & Top-1 & Top-5 & Top-1 & Top-5 & Top-1 & Top-5 \\ \hline
 \multicolumn{2}{|l|}{Full-precision (FP)} & 61.8 & 83.5 & 70.3 & 89.5 & 73.8 & 91.4 \\ \hline
 
XNOR-Net  & 1 / 1 & 44.2 & 69.2 & 51.2 & 73.2 & - & - \\ 
Bi-Real Net  & 1 / 1 & - & - & 56.4 & 79.5 & 62.2 & 83.9 \\
BONN & 1 / 1 & - & - & {59.3} & \textbf{81.6} & - & - \\
\red{RBCN} & \red{1 / 1} & - & - & \red{\textbf{59.5}} & \red{\textbf{81.6}} & - & - \\
\textbf{LQW + CAQ} & 1 / 1 & \textbf{48.3} & \textbf{72.7} & {56.9} & {79.8} & \textbf{62.4} & \textbf{84.0} \\ \hline

DoReFa-Net  & 1 / 2 & 49.8 & - & 53.4 & - & - & - \\ 
TBN  & 1 / 2 & 49.7 & 74.2 & 55.6 & 79.0 & 58.2 & 81.0 \\
LQ-Nets  & 1 / 2 & 55.7 & 78.8 & 62.6 & {84.3} & \textbf{66.6} & 86.9 \\ 
\textbf{LQW + CAQ} & 1 / 2 & \textbf{56.4} & \textbf{79.3} & \textbf{63.1} & \textbf{84.7} & {66.5} & \textbf{87.0} \\\hline
 
DoReFa-Net & 2 / 2 & 48.3 & 71.6 & {57.6} & {80.8} & - & -   \\ 
HWGQ  & 2 / 2 & 52.7 & 76.3 & 59.6 & 82.2 & - & -  \\
LQ-Nets  & 2 / 2 & 57.4 & 80.1 & {64.9} & {85.9} & 69.8 & 89.1 \\
DSQ & 2 / 2 & - & - & 65.2 & - & 70.0 & - \\
QIL  & 2 / 2 & \textbf{58.1} &  - & 65.7 & - & \textbf{70.6} & - \\
\textbf{LQW + CAQ} &  2 / 2 & {57.8} & \textbf{80.4} & \textbf{66.0} & \textbf{86.2} & {70.5} & \textbf{89.4}  \\ \hline
Group-Net  & 4 bases$^\ddagger$ & - & - & 66.3 & 86.6 & - & -  \\
  \hline
 
\end{tabular}
\begin{tablenotes}
\footnotesize
\item $^\ddagger$ The complexity of 4-base Group-Net for a convolution is comparable to ours at $K_w$/$K_a=2/2$.
\item ``-'' indicates that the results are not provided.
\end{tablenotes}
\end{threeparttable}
\vspace{-5pt}
\caption{Top-1/Top-5 classification accuracy (\%) of AlexNet and ResNet-18/34 on ImageNet dataset under different bit-widths.  }
\label{tab:compare_imagenet}
\vspace{-9pt}
\end{table*}

\subsection{Ablation Analysis}
In this section, we first conduct experiments to analyze the effect of our proposals, i.e., \textbf{\textit{(i)}} directly learning the quantized weights (LQW) and \textbf{\textit{(ii)}} the channel-wise average quantizer (CAQ) for activations. For easy comparison, we consider the learned quantizer proposed by \citeauthor{LQNets}, which is used for both weights and activations, as the baseline. Table \ref{tb:ablation_study} shows the classification accuracy of AlexNet on CIFAR-100 with different combinations of baseline method and our proposals for $K_w/K_a\hspace{-1pt}=\hspace{-1pt}2/2$. 
The experimental results demonstrate the effectiveness of our proposals, i.e., LQW and CAQ, in training networks with low bit-width weights and activations. Additionally, LQW helps achieve a slightly higher improvement gain compared to CAQ ($+0.1\%$).

We additionally conduct an experiment on CIFAR-100 using AlexNet with different numbers of CAQ iteration $T$. We repeat the experiment 5 times and report the results (mean$\pm$std) in Table \ref{tab:num_iter}. We observer that using $T = 2,3$ does not result in any significant difference on Top-1 accuracy compared to $T = 1$. 
This is mainly because, for each mini-batch, CAQ starts from proper initial quantization basis values, which are the good results of the last mini-batch.

\subsection{Comparison With State of The Art}
In this section, we compare the performance of our proposed method with
existing methods including 
XNOR-Net \cite{xnor_net}, DoReFa-Net \cite{dorefa}, 
HWQG \cite{hwgq}, PQ+TS+Guided \cite{guidedQuantize},
TBN \cite{tbn}, LQ-Nets \cite{LQNets}, Bi-Real Net \cite{bi-real}, QIL \cite{QIL},
DSQ \cite{dsq}, BONN \cite{bonn}, RCBN \cite{RBCN}, 
and Group-Net \cite{groupnet}. 
Noting that, our method is generic and can be used with any bit-width and architecture.

\paragraph{Comparison on CIFAR-100.}
Table \ref{tb:compared_cifar100} presents the top-1/top-5 classification accuracy on CIFAR-100 dataset of different network quantization methods on AlexNet, ResNet-18, and MobileNetV2. In all compared settings for AlexNet and ResNet-18, our method consistently outperforms the state-of-the-art method LQ-Nets by large margins, i.e., $\ge 0.7\%$ for $K_w/K_a=1/2$ and $\ge 0.6\%$ for $K_w/K_a=2/2$ in term of top-1 accuracy. 
\red{Regarding the efficient architectures MobileNetV2, at bit-widths of $K_w/K_a=3/3$, there is only a small accuracy drop compared with FP model.
Even at lower bit-width $K_w/K_a=2/2$, $K_w/K_a=1/2$, our proposed method still achieves good performance for MobileNetV2, which is already very efficient. This confirms the effectiveness of our proposed method.}
Furthermore, we can observe from Figure \ref{fig:cross_entropy_loss} that the training loss on CIFAR-100 of LQ-Nets with ResNet-18 architecture is unable to approach $0$ as our proposed method does. LQ-Nets also suffers from unstable test error (Figure \ref{fig:test_error}).
This result is potentially due to the mismatch in the training process of LQ-Nets, as the gradient descent does not directly update the quantized weights. Consequently, the quantized weights are unable to obtain the desirable changes.
This demonstrates the necessity of direct updating the quantized weights using gradients as our proposed method can do.

\paragraph{Comparison on ImageNet.}
The classification results on ImageNet dataset for AlexNet and ResNet-18/34 are presented in Table \ref{tab:compare_imagenet}. 
Our proposed method (LQW + CAQ) outperforms the compared methods, e.g., DoReFa-Net, TBN, HORQ, HWGQ, DSQ, in the majority of settings for both AlexNet and ResNet-18/34. 
Our method outperforms LQ-Nets in term of Top-1 accuracy by clear margins in the majority of experiments (i.e., $\ge 0.4\%$);
except for ResNet-34 with $K_w/K_a = 1/2$, where our method achieves comparable performance with LQ-Nets.
For $K_w/K_a = 1/1$, our method gains 0.5\% and 0.2\% Top-1 improvement over Bi-Real Net for ResNet-18 and ResNet-34 respectively. Our method still under-performs BONN \red{and RBCN} for ResNet-18. However, BONN is designed specifically for 1-bit CNN, it is nontrivial to adopt this method to higher bit-widths for better trade-off between the computational speed and accuracy, as can be done easily in our proposed method. \red{Besides, in BONN, in addition to quantization loss and the essential cross-entropy loss, BONN also has an additional feature loss 
to enhance the intra-class compactness which improves the classification accuracy. For our work, because we aimed to learn the quantized network without any modification in the network architecture, in addition to cross-entropy, we only introduce the quantization loss. 
{Regarding RBCN, this method requires joint training of a FP model, a quantized model, and discriminators for distillation from the FP model to the a quantized model. Consequently, this method requires significantly high computational cost and memory.}}
In comparison to QIL, our method achieves comparable performance for $K_w/K_a=2/2$. Specifically, our method has 0.3\% Top-1 accuracy lower than QIL for AlexNet, while achieves 0.3\% higher than QIL for ResNet-18. For ResNet-34, the accuracy of our method is marginally lower than QIL's (i.e., -0.1\% top-1 accuracy). We note that QIL usually achieves the best performance with trainable but complex quantizer functions. It is unsure if the computational complexity of QIL during inference can be linearly proportional to the bit-widths of weights and activations (no efficiency analysis is provided by \citeauthor{QIL}). In our work, we carefully design the method to 
ensure the complexity linearly proportional to the bit-widths. 
Additionally, in comparison with Group-Net (4 bases) on ResNet-18, our method has 0.3\% lower in top-1 accuracy.
Nevertheless, Group-Net requires additional computational costs for the skip-connection in every binary convolution 
and for computing the soft connection between binary groups.

%

\section{Conclusion}
In this paper, we have presented a novel method to train CNN with low bit-width weights and activations. Firstly, we proposed to directly learn the quantized weights with learnable quantization levels using gradient descent, instead of learning to quantize full-precision learned weights as proposed in existing works. 
Secondly, by considering the quantization errors of all activation channels, our proposed quantizer for activations  minimizes the information loss in the majority of channels, instead of being biased toward a few high-variance channels.
The experiment results demonstrate the effectiveness of our proposed method in training low bit-width CNN to achieve the competitive classification accuracy for various network architectures.

\bibliographystyle{named}
\bibliography{ijcai20}

\end{document}